\ifdefined\pdfoutput\pdfoutput=1\fi 
\documentclass{banglab}

\makeatletter
\let\logoheight\BL@rowheight
\newcommand\centeredlogo[2]{\raisebox{\dimexpr(\logoheight-#1\logoheight)/2\relax}{#2}}
\newcommand\logosep{\hspace{0.55\logoheight}{\color{labrule}\rule[-0.02\logoheight]{0.5pt}{1.04\logoheight}}\hspace{0.55\logoheight}}
\newcommand\titlelogos[1]{\renewcommand\BL@lockup[1]{\setlength\logoheight{##1}\mbox{#1}}}
\makeatother

\usepackage{multirow}
\usepackage{makecell}
\usepackage{dsfont}
\usepackage{pgfplots}
\pgfplotsset{compat=1.18}

\newcommand{\method}{\texttt{LatentHarness}}

\newcommand{\titleicon}{\raisebox{-2.4pt}{\includegraphics[height=15pt]{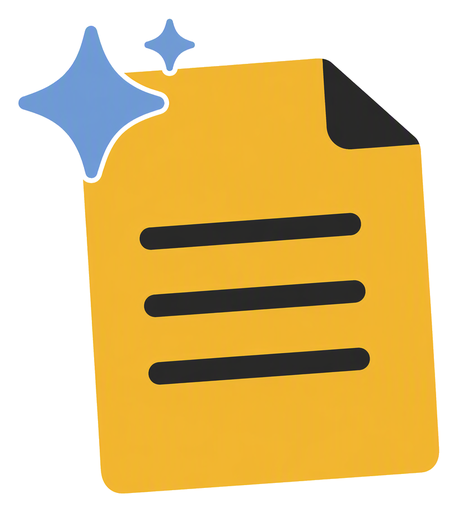}}\,}

\definecolor{r1}{HTML}{fdcdac}
\definecolor{lr}{HTML}{bebada}
\definecolor{l1}{RGB}{189,215,238}
\newcommand{\Mems}[1]{\ensuremath{\mathbf{M}_{#1}}}   

\newcommand{\act}{\ensuremath{a}}                   
\newcommand{\ind}{\ensuremath{\mathds{1}}}            
\newcommand{\Ret}{\ensuremath{R}}                   
\newcommand{\actTHINKop}{\textsc{Think}}
\newcommand{\actRECALL}{\textsc{Recall}}
\newcommand{\nloopmax}{\ensuremath{N}}
\newcommand{\Wlm}{\ensuremath{W_{\mathrm{LM}}}}
\newcommand{\Gain}{\ensuremath{\mathcal{G}}}           
\newcommand{\Llatent}{\ensuremath{\mathcal{L}_{\mathrm{latent}}}}
\newcommand{\Lact}{\ensuremath{\mathcal{L}_{\mathrm{act}}}}
\newcommand{\Lcpd}{\ensuremath{\mathcal{L}_{\mathrm{CPD}}}}
\newcommand{\Lref}{\ensuremath{\mathcal{L}_{\mathrm{ref}}}}
\newcommand{\Lmem}{\ensuremath{\mathcal{L}_{\mathrm{mem}}}}
\newcommand{\Fblk}{\ensuremath{F_\theta}}              
\newcommand{\actEXIT}{\textsc{Exit}}
\newcommand{\polr}{\ensuremath{\pi_\phi}}              
\newcommand{\polstar}{\ensuremath{\pi^{*}}}            
\newcommand{\traj}{\ensuremath{\tau}}
\newcommand{\temp}{\ensuremath{\zeta}}
\newcommand{\Rtask}{\ensuremath{R_{\mathrm{task}}}}
\newcommand{\sstateml}{\ensuremath{s}}                 
\newcommand{\grp}{\ensuremath{G}}
\newcommand{\adv}{\ensuremath{\hat A}}

\definecolor{oursrow}{HTML}{FCF0DC}                    
\definecolor{gaingreen}{HTML}{1B7F3B}
\definecolor{lossred}{HTML}{B03A2E}
\newcommand{\drop}[1]{{\color{lossred}\scriptsize(#1)}}
\newcommand{\rise}[1]{{\color{gaingreen}\scriptsize(#1)}}
\newcommand{\bestin}[2]{\textbf{#1}\,{\tiny\color{gaingreen}(#2)}}   
\newcommand{\famrow}[2]{\multicolumn{#1}{l}{\footnotesize\textit{#2}}}
\newcommand{\na}{\textendash}

\title{\texorpdfstring{\protect\titleicon}{}LatentHarness: Learning Latent Actions\\ for Memory and Reasoning via\\ Counterfactual Policy Distillation}
\titlelogos{%
  \centeredlogo{1.18}{\includegraphics[height=1.18\logoheight]{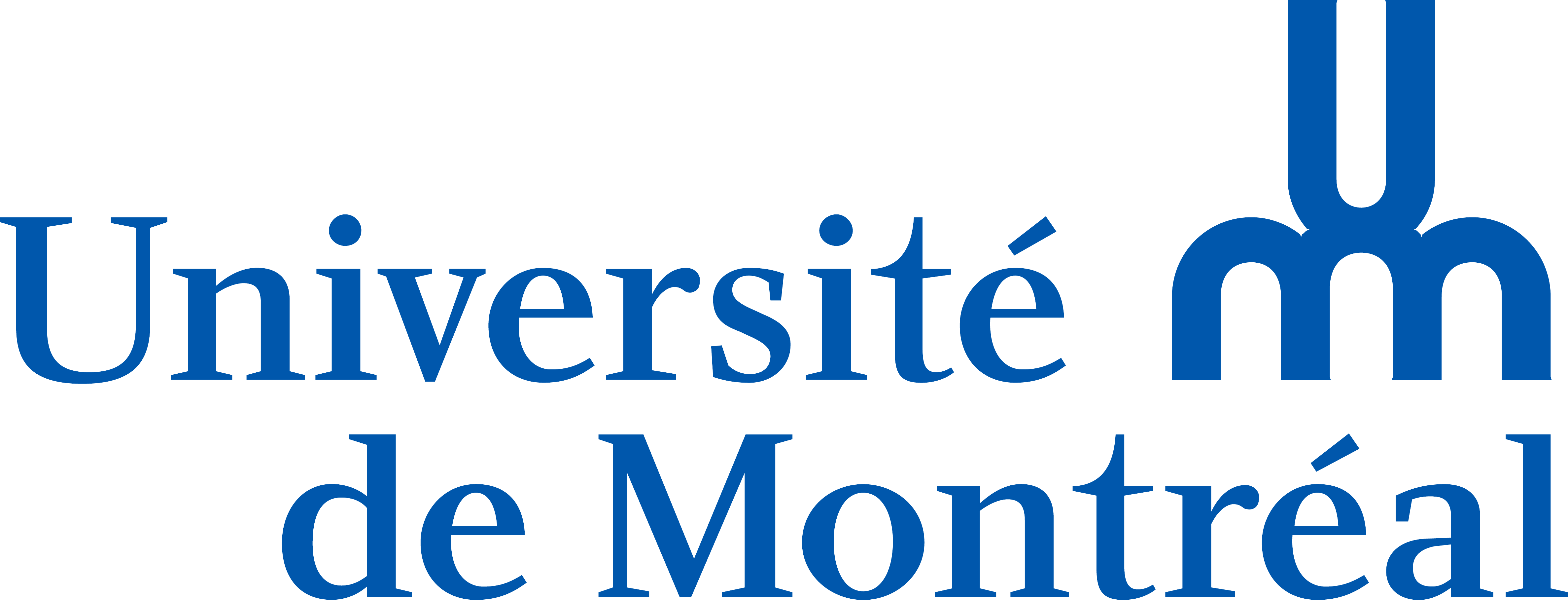}}\logosep
  \centeredlogo{1.0}{\includegraphics[height=\logoheight]{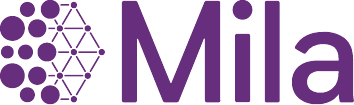}}}

\author[1,2]{Xiaoqiang Wang}
\author[1,2]{Suyuchen Wang}
\author[1,2]{Bang Liu}

\affiliation[1]{Universit\'e de Montr\'eal}
\affiliation[2]{Mila -- Quebec AI Institute}

\abstract{%
Long-context reasoning faces two complementary bottlenecks: retaining evidence across long inputs and sustaining computation across many reasoning steps. Existing approaches largely address them separately, with external memory extending access to distant evidence and latent reasoning compressing multi-step computation. We introduce \textbf{\method{}}, which unifies memory access and latent reasoning as sequential latent action selection. At each internal step, the model chooses \actTHINKop{} for further computation, \actRECALL{} from a fast-weight memory of input evidence and intermediate reasoning states, or \actEXIT{} to emit the next token. We train this policy with \emph{counterfactual policy distillation}, which branches every action for one step and scores its effect on the emitted token. These gains teach the policy when memory is more useful than further reasoning, while gradients through counterfactual recall teach which intermediate states should be retained in memory for future use. Across six general and long-context reasoning benchmarks, \method{} at 1.4B improves on the strongest baselines by 2.8\% and 10.0\% relative, respectively, and runs 5.9$\times$ faster than the strongest long-context baseline.
}

\date{\today}
\correspondence{\email{bang.liu@umontreal.ca}}

\teaser{\centering\includegraphics[width=\linewidth]{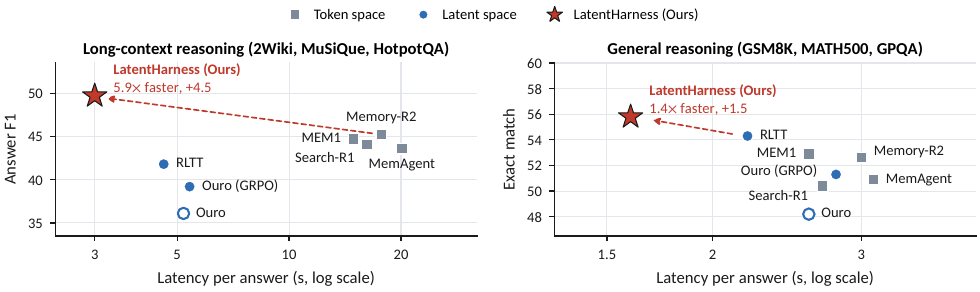}}{Accuracy versus latency of \method{} on general and long-context reasoning with Ouro-1.4B-Thinking. The dashed arrow points from the strongest baseline to ours.\label{fig:teaser}}

\begin{document}

\maketitle

\section{Introduction}
\label{sec:introduction}

Large language models~\citep{achiam2023gpt,yang2025qwen3,guo2025deepseek,wang-etal-2024-fac2e} increasingly act as agents over long horizons~\citep{yao2022react,yang2024sweagent,wang2024openhands,wang2025oscar,li2025webthinker,luo2025large,liu2025advances,ding2026combodied,shi2026evolving}, so their reasoning becomes a long-context problem as trajectories accumulate observations, tool outputs, and intermediate reasoning states~\citep{wu2025resum,ye2025agentfold,lu2025summarization}. This problem involves two distinct forms of length. First, relevant evidence may be distributed across long inputs~\citep{lee2025oolong,bai2024longbenchv2,hsieh2024ruler} or multi-turn interactions~\citep{maharana2024evaluating,wu2024longmemeval}. Second, solving the task may require many intermediate reasoning steps~\citep{wei2022chain,guo2025deepseek,team2025kimi}, and these steps may be interleaved with retrieval of the necessary evidence~\citep{li2025searcho1,jin2025searchr1}.

\begin{figure*}[t]
  \centering
  \includegraphics[width=0.9\linewidth]{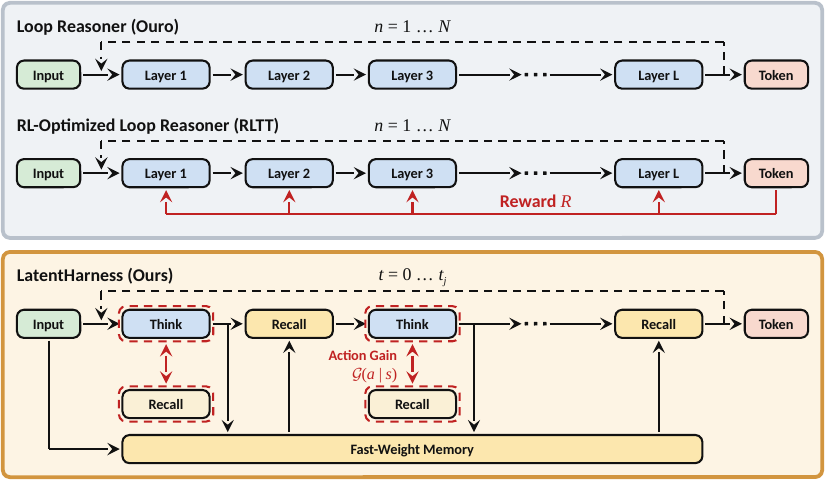}
  \caption{Comparison of Ouro~\citep{ouro2025}, a looped reasoner that recurs over one weight-tied block, RLTT~\citep{rltt2026}, the same loop trained with an outcome reward, and \method{}, which interleaves latent reasoning and memory, with \actTHINKop{} writing the fast-weight memory that \actRECALL{} reads, and learns each latent action from its action gain $\Gain(a\mid\sstateml)$ of \cref{eq:gain}.}
  \label{fig:latent-actions}
\end{figure*}

Existing approaches largely address these two forms of length separately. External-memory agents~\citep{memoryr2_2026,yu2026agentic,mem1_2025,yu2025memagent} retain and retrieve distant evidence across a trajectory, which extends access beyond the current context window. Latent-reasoning methods~\citep{hao2024training,shen2025codi,system15_2025,ouro2025} instead move long intermediate computation from decoded text into internal representations. More recent methods move toward learned memory management by training models to decide when to compress the context~\citep{mem1_2025,wu2025resum,sun2025contextfolding}, and when to read or write stored entries~\citep{yan2025memory,zhang2026memrl,zhang2026deltamem,dong2026memarbiter,li2026ember}. However, these policies still rely either on memory that remains external to the model and is read back through the context window, or on compression rules fixed in advance. This limitation motivates a more direct question: \emph{can memory access and reasoning be unified natively within the model's latent computation?}

As illustrated in \cref{fig:latent-actions}, \method{} builds on the looped computation of latent-reasoning models and unifies memory access and reasoning as sequential latent action selection. At each internal step, a latent policy chooses \actTHINKop{} to continue computation, \actRECALL{} to access persistent memory, or \actEXIT{} to emit the next token. These actions operate over a shared fast-weight associative memory that stores both input evidence and intermediate reasoning states, allowing \actRECALL{} to recover distant evidence or reuse prior computation while \actTHINKop{} deepens reasoning.

Learning such a policy requires assigning credit to individual latent decisions beyond the final task reward. Dense objectives such as RLTT~\citep{rltt2026} supervise what each latent state predicts, but sampled action credit evaluates only the chosen action and vanishes when rollouts stop splitting across actions. We introduce \emph{counterfactual policy distillation}, which branches every allowed latent action, including \actEXIT{}, for one step at each latent state and measures the resulting change in the emitted token's log-probability. These counterfactual branches serve two roles. As \emph{state-level action credit}, their gains teach the policy whether a memory read, further reasoning, or exiting yields the largest advantage-signed, cost-charged one-step gain. As \emph{gain-credited memory writing}, their recall changes, differentiated through the fast-weight memory on positive-advantage trajectories, credit each earlier \actTHINKop{} write by how much it raises or lowers later recalls, which teaches the write gate which derived states to write for later reuse.

\begin{figure*}[t]
  \centering
  \includegraphics[width=0.9\linewidth]{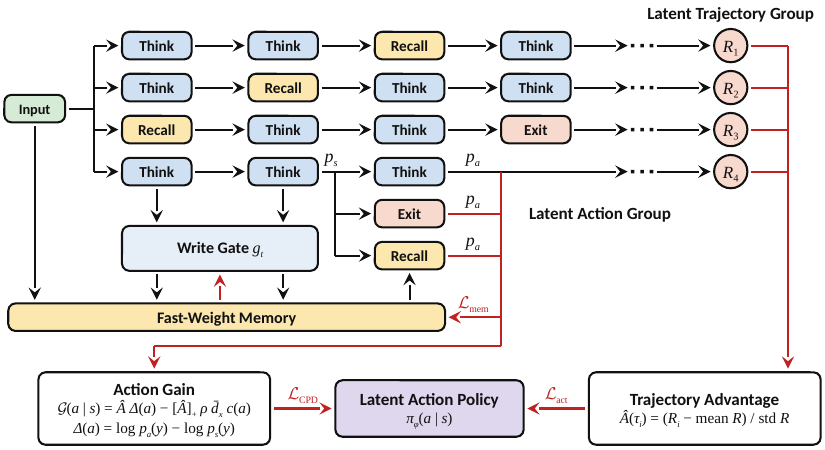}
  \caption{Illustration of counterfactual policy distillation (CPD) in \method{}. The $\grp$ latent trajectories sampled from one input form the latent trajectory group, and their rewards determine the trajectory advantage $\adv(\tau)$. The one-step branches at each latent state form the latent action group, and scoring their decodes $p_a$ against $p_s$ yields the action gain distilled into the latent action policy $\polr$.}
  \label{fig:training}
\end{figure*}

\method{} improves both reasoning quality and efficiency across three general and three long-context reasoning suites. It raises the six-suite average over the strongest baseline by 3.9 points at 1.4B and 4.1 points at 2.6B. At 1.4B, it improves on the strongest baseline in each family by 2.8\% relative on general reasoning and 10.0\% on long-context reasoning, while using 55\% of the backbone's full-depth compute. As \cref{fig:teaser} previews, on long-context reasoning \method{} is 5.9$\times$ faster than Memory-R2 and 1.5$\times$ faster than RLTT.

\section{\texorpdfstring{\method{}}{LatentHarness}: Latent Actions for Memory and Reasoning}
\label{sec:latentharness}

\method{} augments looped latent reasoning with memory recall, so a latent state can read missing evidence or reuse an earlier result instead of recomputing it. At every latent state, a latent action policy chooses among three actions. \actTHINKop{} applies the shared block once more and writes its result to a latent memory that also stores the memorized prompt, \actRECALL{} injects a read from this memory into the hidden state, and \actEXIT{} emits the next token. Because trajectory-level rewards credit latent actions and memory writes only coarsely, counterfactual policy distillation (CPD) branches every allowed action for one step at each latent state, as \cref{fig:training} illustrates. The resulting action gains teach the policy which action to take, while the recall changes, differentiated through memory, teach the write gate which results to store.

\subsection{Latent Actions and Memory}
\label{sec:latent-actions-and-memory}

A loop can only recompute from its current hidden state. \method{} therefore augments it with a latent memory, a fixed-size fast-weight matrix~\citep{schlag2021fwp,behrouz2024titans,sun2024learning,behrouz2026s} written by the forward pass, not by gradient descent. It then treats each output token as the outcome of a decision process over the hidden state, memory, and number of \actTHINKop{} steps taken. Formally, given a prompt $x$, the model emits an answer $y=(y_1,\ldots,y_J)$ one position at a time. Within a position, latent steps are indexed by $t=0,1,\ldots$ with the position index suppressed. The state at step $t$ is $\sstateml_t=(h_t,\Mems{t},n_t)$, where $h_t\in\mathbb{R}^{d}$ is the hidden state, $\Mems{t}\in\mathbb{R}^{d\times d}$ is the latent memory, and $n_t$ counts the \actTHINKop{} steps taken at the position. The latent action policy $\polr$ with parameters $\phi$ selects $\act_t\in\{\actTHINKop,\actRECALL,\actEXIT\}$, the deterministic transition $T_{\act_t}$ produces the next state, and \actEXIT{} at step $t_j$ leaves the state unchanged and emits the token at position $j$,
\begin{equation}
\act_t\sim\polr(\cdot\mid\sstateml_t),\qquad \sstateml_{t+1}=T_{\act_t}(\sstateml_t),\qquad y_j\sim p(\cdot\mid h_{t_j}),
\label{eq:step}
\end{equation}
where $p(\cdot\mid h)=\operatorname{softmax}(\Wlm h)$ decodes a hidden state through the output projection $\Wlm$. Each position starts from its input hidden state $h_0$, $n_0=0$, and the previous position's memory $\Mems{0}$.

\paragraph{Think.}
When the prediction requires further computation, \actTHINKop{} applies the shared block once more and stores the result under a key formed from the hidden state before the block, allowing a similar state to reuse it without another block application. Formally, a looped language model~\citep{ouro2025} applies one shared Transformer block $\Fblk$ with backbone parameters $\theta$ up to $\nloopmax$ times at each position, so that depth grows without added parameters, and uses a learned exit. We leave the block's attention over earlier positions implicit. After $n$ applications from the input hidden state $h^{(0)}$, its state $h^{(n)}=\Fblk\circ\cdots\circ\Fblk(h^{(0)})$ decodes at any depth. \method{} replaces the exit with the latent action policy and exposes each block application as \actTHINKop{}. It also writes by the delta rule~\citep{schlag2021fwp,yang2024deltanet} with the normalized key $k_t=W_kh_t/\lVert W_kh_t\rVert$, the value $v_t=W_vh_{t+1}$ of the block output $h_{t+1}$, and the write strength $g_t=\sigma(w^{\top}h_{t+1}+\kappa\,n_t/\nloopmax)$,
\begin{equation}
h_{t+1}=\Fblk(h_t),\qquad n_{t+1}=n_t+1,\qquad \Mems{t+1}=\Mems{t}+g_t\big(v_t-\Mems{t}k_t\big)k_t^{\top},
\label{eq:think}
\end{equation}
where $W_k$, $W_v$, and the gate vector $w$ are learned, $\sigma$ is the sigmoid, and the learned scalar $\kappa\ge0$ favors results derived after more \actTHINKop{} steps. Without recalls, $h_t=h^{(n_t)}$. Before generation, one block application over the prompt writes every prompt token in order into the zero matrix by the same rule at unit write strength, memorizing the prompt for the first position.

\paragraph{Recall.}
When the hidden state needs information already in memory, \actRECALL{} reads it with one matrix-vector product, which is much cheaper than a block application. Formally, with the normalized query $q_t=W_qh_t/\lVert W_qh_t\rVert$ and learned projections $W_q$ and $W_o$, the \actRECALL{} transition is
\begin{equation}
m_t=\Mems{t}q_t,\qquad h_{t+1}=h_t+W_om_t,\qquad \Mems{t+1}=\Mems{t},\quad n_{t+1}=n_t,
\label{eq:recall}
\end{equation}
where the read $m_t$ superposes stored values whose keys resemble $q_t$.

\paragraph{Latent action policy.}
The policy must judge whether a recall would improve the prediction before the read changes the hidden state. It therefore observes the hidden state and a probe of the pending read. Computing this probe is not a latent step, whereas selecting \actRECALL{} is. Formally, let $u_t=(\lVert m_t\rVert_2,\cos(m_t,h_t))$ be the probe of the read $m_t$ in \cref{eq:recall}. The learned linear head $W_\pi$ defines $\polr(a\mid\sstateml_t)=\operatorname{softmax}(W_\pi[h_t;u_t])$ over the three actions, where $[h_t;u_t]$ denotes concatenation. The selected action updates the hidden state as
\begin{equation}
h_{t+1}=\underbrace{\ind[\act_t{=}\actTHINKop]\,\Fblk(h_t)}_{\textstyle\text{compute}}+\underbrace{\ind[\act_t{=}\actRECALL]\,\big(h_t+W_om_t\big)}_{\textstyle\text{recall}}+\underbrace{\ind[\act_t{=}\actEXIT]\,h_t}_{\textstyle\text{emit }y_j\sim p(\cdot\mid h_t)},
\label{eq:transition}
\end{equation}
where $\ind[\cdot]$ is the indicator function. \actTHINKop{} is masked when $n_t=\nloopmax$, and \actRECALL{} is masked after $\nloopmax$ recalls at the position. An implicit per-position counter tracks recalls and is read by the admissible set $\mathcal{A}(\sstateml_t)$ of unmasked actions. \actEXIT{} is forced when both alternatives are masked. $\polr$ is renormalized over $\mathcal{A}(\sstateml_t)$, rollouts sample from it, and inference selects the most probable action. No gradient passes through selection, so $\phi=\{W_\pi\}$ learns only through $\log\polr$, while the backbone parameters $\theta$ and memory parameters $\psi=\{W_q,W_o,W_k,W_v,w,\kappa\}$ learn through realized states. The recall branches of \cref{sec:counterfactual-policy-distillation} also train $w$ and $\kappa$.

\subsection{Counterfactual Policy Distillation}
\label{sec:counterfactual-policy-distillation}

\paragraph{Reinforcement learning over latent trajectories.}
We train whole latent trajectories with GRPO on the task reward. Formally, a rollout of \cref{eq:step} over the positions $j=1,\ldots,J$ of a prompt $x$, with the position index restored, produces the latent trajectory $\traj=\big((\sstateml_{j,t},\act_{j,t})_{t=0}^{t_j},\,y_j\big)_{j=1}^{J}$. Because every transition is deterministic given the state and earlier positions, its likelihood factors as
\begin{equation}
\pi(\traj\mid x)=\prod_{j=1}^{J}\Big(\prod_{t=0}^{t_j}\polr(\act_{j,t}\mid\sstateml_{j,t})\Big)\,p(y_j\mid h_{j,t_j}),
\label{eq:traj}
\end{equation}
where $\act_{j,t_j}=\actEXIT$. The reward $\Ret(\traj)=\Rtask(y)$ is the answer's exact match or answer F1, without a length penalty, since a penalty could rank a short failure above a longer success. GRPO~\citep{shao2024grpo} samples $\traj_1,\ldots,\traj_{\grp}$ per prompt, as \cref{fig:training} shows, and normalizes their rewards by the group mean and standard deviation into advantages $\adv(\traj_i)$. The loss $-\adv(\traj)\log\pi(\traj\mid x)$, whose gradient at the on-policy point equals that of the clipped surrogate of PPO~\citep{schulman2017proximal}, decomposes by \cref{eq:traj} into the sampled-action loss $\Lact$ and a token term $-\adv(\traj)\sum_j\log p(y_j\mid h_{j,t_j})$, with each token and latent-action ratio clipped separately. Because the token term supervises only exit states, we follow RLTT~\citep{rltt2026} and replace it with the dense latent loss $\Llatent$, yielding
\begin{equation}
\Lact=-\adv(\traj)\sum_{j}\sum_{t}\log\polr(\act_{j,t}\mid\sstateml_{j,t}),\qquad
\Llatent=-\adv(\traj)\sum_{j}\sum_{t}\omega_{n_{j,t}}\log p(y_j\mid h_{j,t}),
\label{eq:latent}
\end{equation}
where the depth weights $\omega_n$ follow RLTT, are shared by states that \actRECALL{} reaches at the same \actTHINKop{} count, and are normalized over the realized states of each position. Two decisions still receive only coarse credit. $\Lact$ assigns one advantage to every latent action in a trajectory, and the write gate, as a deterministic part of the transition, receives gradients only through predictions downstream of executed recalls, since the probe $u$ feeds only the policy losses, which update only $\phi$.

\paragraph{State-level action credit.}
At one latent state, the shared advantage of $\Lact$ provides credit that fades. Its gradient on the logit of an action taken by a fraction $\hat p$ of the sampled continuations scales with $\hat p(1-\hat p)$, so it vanishes when they do not split and fades as the policy commits. CPD instead compares every admissible action from the same state using the one-step change it makes to the emitted token's log-probability, scaled by trajectory quality and charged an action cost for its latent length on positive-advantage trajectories. Formally, for each $a\in\mathcal{A}(\sstateml)$, including \actEXIT{}, the one-step change and the action gain are
\begin{equation}
\Delta(a\mid\sstateml)=\log p_{T_a(\sstateml)}(y)-\log p_{\sstateml}(y),\quad
\Gain(a\mid\sstateml)=\adv(\traj)\big(\Delta(a\mid\sstateml)-\ind[\adv(\traj)>0]\,\rho\,\bar d_x\,c(a)\big),
\label{eq:gain}
\end{equation}
where $y$ is the token that $\traj$ emits at this position, $p_{\sstateml}(y)$ and $p_{T_a(\sstateml)}(y)$ are its probabilities decoded before and after $a$, $\rho$ is the cost weight, and $\bar d_x$ is the median $|\Delta|$ over the \actTHINKop{} and \actRECALL{} branches of the prompt's positive-advantage rollouts. The latent lengths are $c(\actTHINKop)=1$, $c(\actRECALL)=c_R<1$, and $c(\actEXIT)=0$. Recall has a shorter latent length because it injects an existing read and applies no block. Because $T_{\actEXIT}$ is the identity, \actTHINKop{} or \actRECALL{} outscores \actEXIT{} on a positive-advantage trajectory exactly when its $\Delta$ exceeds its cost. Without the indicator, a negative advantage would reward latent length.

The gains define a counterfactual teacher over $\mathcal{A}(\sstateml)$ as the Boltzmann policy $\polstar(a\mid\sstateml)\propto\exp\big(\Gain(a\mid\sstateml)/\temp\big)$. CPD distills this teacher into the policy through policy distillation~\citep{rusu2015policy},
\begin{equation}
\Lcpd=\frac{1}{|\mathcal{S}_{\traj}|}\sum_{\sstateml\in\mathcal{S}_{\traj}} D_{\mathrm{KL}}\big(\polstar(\cdot\mid\sstateml)\,\Vert\,\polr(\cdot\mid\sstateml)\big),
\label{eq:cpd}
\end{equation}
where $\mathcal{S}_{\traj}$ is the set of latent states of $\traj$, the gains are held under stop-gradient, and $\Lcpd$ is averaged over the group. The temperature $\temp$ controls how sharply the teacher favors higher gains, approaching the hard label $\arg\max_a\Gain(a\mid\sstateml)$ as $\temp\to0$ and the uniform distribution over $\mathcal{A}(\sstateml)$ as $\temp\to\infty$. Because the teacher is fixed, $\Lcpd$ is cross-entropy to a soft label, with gradient $\polr(a\mid\sstateml)-\polstar(a\mid\sstateml)$ on the logit of $a$, so its credit does not fade as the policy commits. Each branch ends after one action and one decode. An untaken branch is never continued and does not commit its write, while the taken branch reuses its next state. The gains therefore provide immediate state-level credit at every latent state, while $\Lact$ carries the downstream credit of the sampled path.

\paragraph{Gain-credited memory writing.}
A write helps only when a later recall reads it and raises the emitted token's probability. Standard delta-rule training therefore credits the gate only downstream of executed recalls, which are rare early in training. CPD provides denser credit through recall branches because every state where \actRECALL{} is admissible measures $\Delta(\actRECALL\mid\sstateml)$, whether or not the policy recalls there. This change depends on every earlier gate through $\Mems{}$. Formally, we train the gate to increase the recall changes of positive-advantage trajectories,
\begin{equation}
\Lmem=-\frac{[\adv(\traj)]_+}{|\mathcal{S}_{\traj}|}\sum_{\sstateml\in\mathcal{S}_{\traj}}\Delta(\actRECALL\mid\sstateml),
\label{eq:mem}
\end{equation}
where $[\adv(\traj)]_+$ is the positive part of $\adv(\traj)$ and masked states contribute zero. $\Lmem$ updates only the gate parameters $w$ and $\kappa$, with every other quantity held fixed, so the token losses still train the keys and values. Restricting this signal to positive advantages is a design choice, since negative weights could make the gate lower recall changes by storing noise instead of correcting a harmful write.

The credit assigned to a write separates into whether its content improves a later prediction and whether the later query still addresses it after intervening writes. Formally, number the latent states of $\traj$ by $r=0,1,\ldots$ across positions. A realized \actTHINKop{} write at step $r$ changes the recall change at a later state $\sstateml_{r'}$ where \actRECALL{} is admissible at the rate
\begin{equation}
\frac{\partial\Delta(\actRECALL\mid\sstateml_{r'})}{\partial g_r}
=\underbrace{\big\langle\nabla_{m}\log p_{T_{\actRECALL}(\sstateml_{r'})}(y_{r'}),\;v_r-\Mems{r}k_r\big\rangle}_{\textstyle\text{content}}\;
\underbrace{k_r^{\top}P_{r+1}\cdots P_{r'-1}\,q_{r'}}_{\textstyle\text{addressing}},
\label{eq:write-credit}
\end{equation}
where $y_{r'}$ is the token emitted at the position of $\sstateml_{r'}$, the gradient is evaluated at the branch read $m=\Mems{r'}q_{r'}$, $P_\ell=I-g_\ell k_\ell k_\ell^{\top}$ at each intervening \actTHINKop{} step and $P_\ell=I$ otherwise, and the empty product is $I$. After scaling by $[\adv(\traj)]_+/|\mathcal{S}_{\traj}|$, a write receives the sum of these products over later admissible states. Its credit is exactly zero when $v_r=\Mems{r}k_r$ or when its addressing factor vanishes at every later admissible state. A later state contributes negatively when the addressed change lowers its prediction. Because hidden states are held fixed, \cref{eq:write-credit} is a partial derivative along the realized trajectory. It ignores how a gate would alter later keys, values, gates, queries, and actions, but requires only one backward pass through branches CPD already computes.

\paragraph{Final objective.}
The objective is $\mathcal{L}=\Llatent+\beta\Lact+\gamma\Lcpd+\lambda\Lmem+\eta\Lref$, where $\beta$, $\gamma$, $\lambda$, and $\eta$ are loss weights and $\Lref$ is the standard KL regularizer~\citep{shao2024grpo} from the token distribution at each emitting state to that of the frozen full-depth backbone. $\Llatent$ and $\Lref$ train the backbone $\theta$ and the memory maps $\psi$ through the realized states, $\Lact$ and $\Lcpd$ train the policy $\phi$, and $\Lmem$ trains the write gate $w$ and $\kappa$. We anneal $\gamma$ linearly from one to a tenth over the first 1.5k steps, so the teacher shapes the policy early and later only prevents commitment.

\section{Experiments}
\label{sec:experiments}

\begin{table*}[t]
  \centering
  \caption{Quantitative results of \method{} with Ouro-1.4B-Thinking and Ouro-2.6B-Thinking. MATH, GPQA, 2WQA, MSQ, and HQA denote MATH500, GPQA-Diamond, 2WikiMultihopQA, MuSiQue, and HotpotQA, $^\dagger$ marks an out-of-distribution suite, and Lat.\ is wall-clock seconds per answer. \textbf{Bold}, \underline{underline}, and {\color{gaingreen}green} mark the best, the second best, and the margin over the second best. Ouro is the released backbone at full depth.}
  \label{tab:main-results}
\small
\setlength{\tabcolsep}{2.7pt}
\begin{tabular}{@{}l ccc ccc ccc ccc@{}}
\toprule
& \multicolumn{6}{c}{General reasoning} & \multicolumn{6}{c}{Long-context reasoning} \\
\cmidrule(lr){2-7}\cmidrule(lr){8-13}
Method & GSM8K & MATH & GPQA$^\dagger$ & Avg. & FLOPs & Lat. & 2WQA$^\dagger$ & MSQ$^\dagger$ & HQA & Avg. & FLOPs & Lat. \\
\midrule
\multicolumn{13}{@{}l}{\textbf{Ouro-1.4B-Thinking}}\\
Ouro                     & 74.6 & 42.8 & 27.3 & 48.2 & 100\% & 2.6 & 41.2 & 19.7 & 47.5 & 36.1 & 100\% & 5.2 \\
Ouro (GRPO)              & 78.1 & 46.3 & 29.4 & 51.3 & 106\% & 2.8 & 44.8 & 22.1 & 50.6 & 39.2 & 108\% & 5.4 \\
\famrow{13}{Token space}\\
\,+ Search-R1        & 77.4 & 45.6 & 28.3 & 50.4 & 133\% & 2.7 & 50.4 & 27.1 & 54.9 & 44.1 & 339\% & 16.2 \\
\,+ MemAgent         & 78.0 & 45.7 & 29.0 & 50.9 & 130\% & 3.1 & 49.7 & 26.8 & 54.3 & 43.6 & 454\% & 20.1 \\
\,+ MEM1             & 79.6 & 48.1 & 31.0 & 52.9 & 121\% & 2.6 & 50.9 & 27.8 & 55.5 & 44.7 & 307\% & 14.9 \\
\,+ Memory-R2        & 79.4 & 47.6 & 30.8 & 52.6 & 125\% & 3.0 & \second{51.3} & \second{28.4} & \second{55.9} & \second{45.2} & 391\% & 17.7 \\
\famrow{13}{Latent space}\\
RLTT                     & \second{81.5} & \second{49.6} & \second{31.8} & \second{54.3} & \second{86\%} & \second{2.2} & 47.6 & 24.5 & 53.2 & 41.8 & \second{82\%} & \second{4.6} \\
\rowcolor{labtint}
\method{}             & \best{83.2} & \best{51.6} & \best{32.6} & \bestin{55.8}{$+$1.5} & \best{62\%} & \best{1.6} & \best{55.9} & \best{33.1} & \best{60.1} & \bestin{49.7}{$+$4.5} & \best{48\%} & \best{3.0} \\
\midrule
\multicolumn{13}{@{}l}{\textbf{Ouro-2.6B-Thinking}}\\
Ouro                     & 79.3 & 48.5 & 31.6 & 53.1 & 100\% & 4.4 & 45.9 & 23.4 & 51.8 & 40.4 & 100\% & 8.8 \\
Ouro (GRPO)              & 82.7 & 52.0 & 33.9 & 56.2 & 105\% & 4.7 & 49.2 & 26.1 & 55.0 & 43.4 & 107\% & 9.3 \\
\famrow{13}{Token space}\\
\,+ Search-R1        & 82.1 & 51.0 & 32.8 & 55.3 & 131\% & 4.5 & 54.6 & 31.0 & 59.0 & 48.2 & 335\% & 27.1 \\
\,+ MemAgent         & 82.6 & 51.5 & 33.6 & 55.9 & 128\% & 5.2 & 54.1 & 30.7 & 58.6 & 47.8 & 448\% & 33.6 \\
\,+ MEM1             & 84.3 & 53.5 & 35.3 & 57.7 & 120\% & 4.4 & 55.0 & 31.6 & 59.4 & 48.7 & 304\% & 24.6 \\
\,+ Memory-R2        & 84.0 & 53.3 & 35.2 & 57.5 & 123\% & 4.9 & \second{55.6} & \second{32.2} & \second{60.1} & \second{49.3} & 387\% & 29.5 \\
\famrow{13}{Latent space}\\
RLTT                     & \second{85.4} & \second{55.3} & \second{36.2} & \second{59.0} & \second{87\%} & \second{3.7} & 52.0 & 28.3 & 57.4 & 45.9 & \second{83\%} & \second{7.7} \\
\rowcolor{labtint}
\method{}             & \best{87.3} & \best{57.2} & \best{38.3} & \bestin{60.9}{$+$1.9} & \best{64\%} & \best{2.6} & \best{60.3} & \best{37.6} & \best{64.2} & \bestin{54.0}{$+$4.7} & \best{50\%} & \best{5.0} \\
\bottomrule
\end{tabular}

\end{table*}

\paragraph{Backbones.}
We evaluate the two released looped reasoners, Ouro-1.4B-Thinking and Ouro-2.6B-Thinking~\citep{ouro2025}, at their trained recurrent depth $\nloopmax=4$. The shared block $\Fblk$ is the full stack of 24 or 48 layers. The memory maps $\psi$ and latent action policy $\phi$ are the only new parameters, adding 1.8\% and 1.0\% to the respective parameter counts.

\paragraph{Datasets.}
We train one model per backbone on the GSM8K~\citep{cobbe2021gsm8k} and MATH~\citep{hendrycks2021measuring} training splits. Following MemAgent~\citep{yu2025memagent} and Search-R1~\citep{jin2025searchr1}, we also train on HotpotQA~\citep{yang2018hotpotqa} questions with supporting passages embedded among distractors and padded to 4k, 8k, and 16k tokens. We test GSM8K, MATH500~\citep{lightman2023let}, and HotpotQA in distribution. Following RLTT~\citep{rltt2026}, we test GPQA-Diamond~\citep{rein2023gpqa} out of distribution, along with 2WikiMultihopQA~\citep{ho2020wikihop} and MuSiQue~\citep{trivedi2022musique}. Each long-context test set is padded to 4k, 8k, 16k, 32k, and 64k tokens, so the two longest lengths exceed every training input. We report exact match on the general suites and answer F1 on the long-context suites.

\paragraph{Baselines.}
GRPO~\citep{shao2024grpo} trains the Ouro backbone with outcome RL and no memory, and every token-space memory method is built on top of it. Search-R1~\citep{jin2025searchr1} interleaves retrieval and reasoning over the padded context, MemAgent~\citep{yu2025memagent} and Memory-R2~\citep{memoryr2_2026} write and read memory as generated text, and MEM1~\citep{mem1_2025} combines memory and reasoning in one compact token-space state per turn. Latent-space methods include RLTT~\citep{rltt2026}, which gives the loop dense latent credit without memory, and \method{}. Every trained method uses the same backbone, training data, and number of RL steps. We report FLOPs relative to the full-depth backbone and per-answer latency following \citet{system15_2025}.

\subsection{Main Results}
\label{sec:main-results}

\paragraph{\method{} achieves the highest accuracy at both model scales.}
In \cref{tab:main-results}, averaging the two family averages gives six-suite means of 52.8 at 1.4B and 57.5 at 2.6B, exceeding Memory-R2 by 3.9 and 4.1 points, respectively. The gains hold within both task families, consistent with a latent action policy that distinguishes states needing further computation from those missing information.

\paragraph{The largest gains come from supplying missing information on long inputs.}
Relative to RLTT, which uses the same latent loop without memory, \method{} raises the general-reasoning average from 54.3 to 55.8 and the long-context average from 41.8 to 49.7. Its long-context margin grows from 2.3 points at 4k tokens to 14.3 at 64k. On the general suites, the hidden state already carries the evidence, so the policy mainly reallocates \actTHINKop{} steps. On long inputs, \actRECALL{} recovers evidence that further latent computation cannot reconstruct.

\begin{table}[t]
  \centering
  \caption{Ablations of \method{} on action credit, action space, write gate, and action cost. Gen.\ and Long denote family averages, FLOPs denotes the two-family average relative to full depth, and parentheses mark the change from ours. Training compute is 1.1$\times$, 1.2$\times$, and 3.6$\times$ GRPO only for ours, GiGPO groups, and VinePPO.}
  \label{tab:ablations}
\small
\setlength{\tabcolsep}{5pt}
\begin{tabular}{@{}l ccc@{}}
\toprule
Variant & Gen. & Long & FLOPs \\
\midrule
\rowcolor{labtint}
\method{} & 55.8 & 49.7 & 55\% \\
\famrow{4}{Action credit}\\
\quad GRPO only & 52.3\,\drop{$-$3.5} & 44.3\,\drop{$-$5.4} & 71\% \\
\quad GiGPO groups & 52.9\,\drop{$-$2.9} & 45.1\,\drop{$-$4.6} & 68\% \\
\quad VinePPO branches & 54.9\,\drop{$-$0.9} & 48.6\,\drop{$-$1.1} & 58\% \\
\famrow{4}{Action space}\\
\quad Think and Exit & 53.6\,\drop{$-$2.2} & 42.0\,\drop{$-$7.7} & 63\% \\
\quad Always read & 54.1\,\drop{$-$1.7} & 45.3\,\drop{$-$4.4} & 58\% \\
\famrow{4}{Write gate}\\
\quad Token-loss gate & 55.5\,\drop{$-$0.3} & 48.1\,\drop{$-$1.6} & 57\% \\
\quad Write-time gate & 55.4\,\drop{$-$0.4} & 47.6\,\drop{$-$2.1} & 58\% \\
\famrow{4}{Action cost}\\
\quad No cost & 56.1\,\rise{$+$0.3} & 50.0\,\rise{$+$0.3} & 93\% \\
\bottomrule
\end{tabular}

\end{table}

\paragraph{\method{} improves accuracy while reducing inference cost.}
At 1.4B, \method{} averages 55\% of full-depth FLOPs across the two task families, compared with 84\% for RLTT, and answers long-context questions 5.9$\times$ faster than Memory-R2 in \cref{fig:teaser}. In \cref{tab:ablations}, removing the action cost ($\rho=0$) raises FLOPs from 55\% to 93\% while improving each family average by only 0.3 points. Applying the action cost to the action gain instead of the trajectory return suppresses low-gain latent steps without allowing a shorter failure to outrank a longer success.

\subsection{In-Depth Analysis}
\label{sec:in-depth-analysis}
\paragraph{RQ1: Does state-level credit matter?}
We retrain the policy with GRPO only, GiGPO groups, and VinePPO branches on the same data and steps. We compare their family averages and training compute with ours in \cref{tab:ablations}, and track the entropy of $\polr$ and the recall change $\Delta(\actRECALL\mid\sstateml)$ on test rollouts.

\paragraph{One-step counterfactual branches outperform completed branches at a third of the training compute.}
VinePPO branches complete each branched action to the answer~\citep{kazemnejad2024vineppo}, while GiGPO groups pool latent states by output position and action prefix~\citep{feng2025gigpo}. On the long-context suites, \method{} reaches 49.7 at 1.1$\times$ training compute, compared with 48.6 at 3.6$\times$ for VinePPO branches. Every latent state decodes, so a one-step branch scores each action without completing a rollout.

\paragraph{CPD keeps the latent action policy from collapsing because its credit does not depend on sampled exploration.}
Under GRPO only, which drops $\Lcpd$, the entropy of $\polr$ falls to 0.21 nats within 200 steps, whereas \method{} holds it at 0.68, and its $\Delta(\actRECALL\mid\sstateml)$ from \cref{eq:gain} averages 0.36 nats at executed recalls against 0.03 at admissible states where it is not taken. Sampled credit fades as the policy commits, whereas counterfactual branches compare every action.

\paragraph{RQ2: When should the policy recall instead of think?}
We retrain the Think and Exit and Always read action-space ablations, and run RLTT at the full depth $\nloopmax=4$ as All-Think. A one-step \actTHINKop{} probe marks each position's settling step, and we count \actTHINKop{} and \actRECALL{} steps per position by that step and by input length. We measure each All-Think step's I/O cosine and, at every latent state of \method{}, the pending read's norm and cosines grouped by executed action.

\paragraph{Selective \actRECALL{} is necessary on long inputs.}
Always read injects a memory read at every step. Removing \actRECALL{} lowers the long-context average by 7.7 points to 42.0, while Always read remains 4.4 points below \method{}. An always-on read cannot distinguish missing evidence from missing computation, so the policy must decide when to read.

\begin{figure*}[t]
  \centering
  \includegraphics[width=\linewidth]{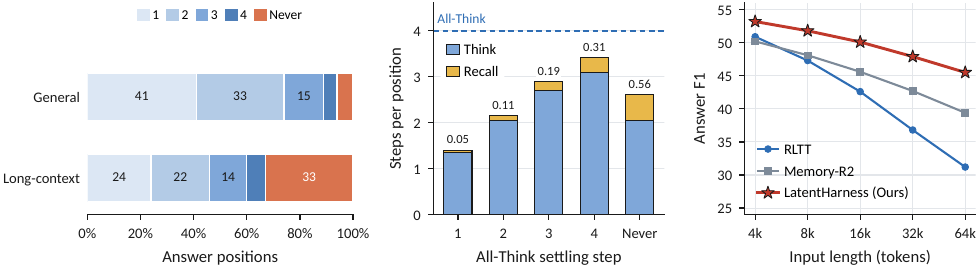}
  \caption{All-Think settling, latent actions, and long-context F1 of \method{}, RLTT, and Memory-R2 on Ouro-1.4B-Thinking. The left, middle, and right panels show positions by All-Think settling step, latent steps per position by settling step, and F1 by input length, respectively.}
  \label{fig:patterns}
\end{figure*}

\paragraph{Full-depth thinking wastes loops yet leaves a third of long-context positions unsettled.}
The left panel of \cref{fig:patterns} shows that 51\% of loops change nothing on the general suites, while 33\% of long-context positions never settle. Further loops are redundant after settlement and cannot recover evidence that the hidden state lacks, which only a read can supply.

\begin{figure*}[t]
  \centering
  \includegraphics[width=\linewidth]{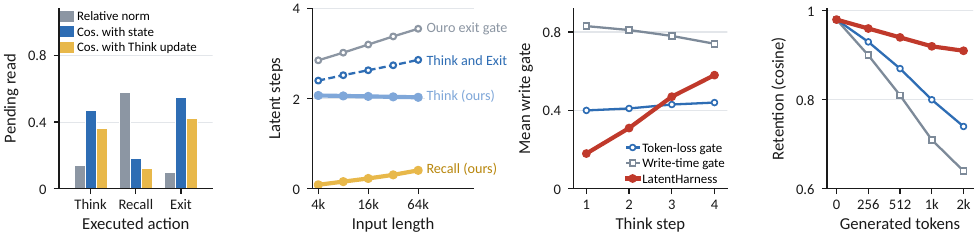}
  \caption{Latent-step dynamics and write gates of \method{} on the Ouro-1.4B-Thinking long-context suites. From left to right, the panels show the pending read at latent states by executed action, latent steps per position by input length, the mean write gate by \actTHINKop{} step under three gate objectives, and evidence retention by generated tokens.}
  \label{fig:latent-dynamics}
\end{figure*}

\paragraph{\actTHINKop{} converges toward the current hidden state, while \actRECALL{} adds a direction unavailable to further \actTHINKop{} steps.}
Following \citet{zhang2026deeper}, a step's I/O cosine is the cosine between its input and output hidden states. Under All-Think, this cosine rises from 0.71 to 0.99 on the general suites but plateaus at 0.91 on the long-context suites. Later loops therefore rewrite either a converged state or one that cannot converge. In the first panel of \cref{fig:latent-dynamics}, the pending read at \actRECALL{} states has a relative norm of 0.58 and a cosine of 0.12 with the pending \actTHINKop{} update, compared with 0.14 and 0.36 at \actTHINKop{} states. This larger, less aligned read adds a component that further computation from the same state cannot produce.

\begin{table*}[t]
  \centering
  \caption{\method{} task scores, latent steps per position, and recall sources on Ouro-1.4B-Thinking, grouped by suite, hop count, and input length. Think and Exit reports the \actTHINKop{} steps of the ablation, while Ouro exit gate reports the loops of the released backbone under its own exit. Derived is the share of recalls that read a \actTHINKop{}-written state, Hit is the share of evidence recalls that reach a supporting passage, and Chance is the share of prompt tokens contained in supporting passages. Score is exact match on the general suites and F1 otherwise. Hop and length rows pool over the other axis, and \na{} marks suites without supporting-fact annotation.}
  \label{tab:per-suite}
\small
\setlength{\tabcolsep}{6pt}
\begin{tabular}{@{}l ccc c c ccc@{}}
\toprule
& \multicolumn{3}{c}{\method{}} & Think and Exit & Ouro exit gate & \multicolumn{3}{c}{Recall sources (\%)} \\
\cmidrule(lr){2-4}\cmidrule(lr){5-5}\cmidrule(lr){6-6}\cmidrule(lr){7-9}
Suite & Think & Recall & Score & Think & Loops & Derived & Hit & Chance \\
\midrule
\famrow{9}{General reasoning}\\
GSM8K    & 2.10 & 0.03 & 83.2 & 2.14 & 2.70 & 39.2 & \na  & \na \\
MATH500  & 2.55 & 0.05 & 51.6 & 2.57 & 3.30 & 35.6 & \na  & \na \\
GPQA     & 2.40 & 0.07 & 32.6 & 2.43 & 3.20 & 33.2 & \na  & \na \\
\famrow{9}{Long-context reasoning}\\
2WQA     & 2.00 & 0.20 & 55.9 & 2.46 & 3.10 & 29.3 & 69.8 & 2.8 \\
MuSiQue  & 2.15 & 0.32 & 33.1 & 2.95 & 3.40 & 34.7 & 65.9 & 2.8 \\
HotpotQA & 2.00 & 0.20 & 60.1 & 2.47 & 3.10 & 27.7 & 72.0 & 2.7 \\
\famrow{9}{MuSiQue by hop count}\\
2-hop    & 2.09 & 0.24 & 42.1 & 2.78 & 3.24 & 29.5 & 74.0 & 2.2 \\
3-hop    & 2.18 & 0.37 & 27.3 & 3.08 & 3.52 & 35.6 & 63.6 & 3.1 \\
4-hop    & 2.27 & 0.50 & 16.4 & 3.25 & 3.67 & 40.8 & 55.6 & 4.1 \\
\famrow{9}{Long-context by input length}\\
4k       & 2.07 & 0.09 & 53.2 & 2.40 & 2.85 & 27.4 & 78.4 & 7.0 \\
8k       & 2.06 & 0.16 & 51.8 & 2.52 & 3.02 & 28.9 & 74.9 & 3.6 \\
16k      & 2.05 & 0.23 & 50.1 & 2.63 & 3.20 & 30.3 & 71.1 & 1.8 \\
32k      & 2.04 & 0.31 & 47.9 & 2.74 & 3.38 & 31.7 & 67.2 & 0.9 \\
64k      & 2.03 & 0.41 & 45.5 & 2.86 & 3.55 & 33.2 & 63.9 & 0.5 \\
\bottomrule
\end{tabular}

\end{table*}

\paragraph{\method{} spends recalls, not extra \actTHINKop{} steps, where evidence is missing.}
In the middle panel of \cref{fig:patterns}, positions settled at step 1 average 0.05 recalls, compared with 0.56 for positions that never settle. In the second panel of \cref{fig:latent-dynamics}, recalls rise from 0.09 to 0.41 per position, while \actTHINKop{} steps remain near 2.0. Across the long-context suites in \cref{tab:per-suite}, \method{} takes 2.0 to 2.2 \actTHINKop{} steps per position, compared with 2.5 to 3.0 for Think and Exit and 3.1 to 3.4 loops under Ouro's exit gate. Each recall therefore replaces further computation. For unsettled positions and long inputs, evidence is missing from the hidden state, which gives \actRECALL{} a larger one-step action gain than another \actTHINKop{}.

\begin{figure*}[t]
  \centering
  \includegraphics[width=\linewidth]{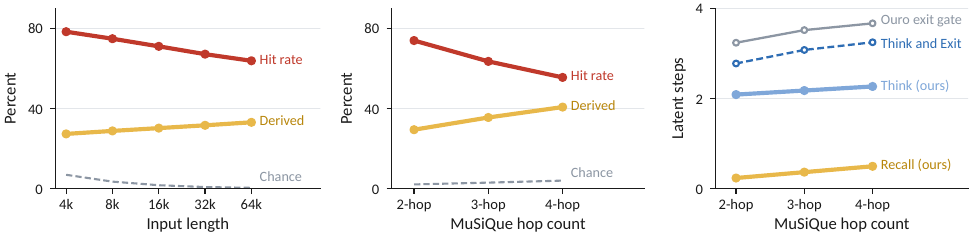}
  \caption{\method{} recall sources and latent steps on the Ouro-1.4B-Thinking long-context suites. The left and middle panels show recall hit rate, derived share, and chance rate by input length and MuSiQue hop count. The right panel shows latent steps per position by hop count.}
  \label{fig:recall-sources}
\end{figure*}

\paragraph{RQ3: What does \actRECALL{} read?}
We split every executed read into memorized-prompt and \actTHINKop{}-written components to label it as derived or evidence, match each evidence recall to its best-matching passage and score hits against supporting-fact annotations, define chance as the share of prompt tokens in supporting passages, report F1 by hop count, and rank passages at the final evidence recall for wrong MuSiQue answers with at least three hops.

\paragraph{\actRECALL{} returns supporting evidence and increasingly reuses derived states.}
In \cref{fig:recall-sources}, evidence recalls achieve a hit rate of 63.9\% at 64k tokens against a chance rate of 0.5\%, and 55.6\% at four hops against 4.1\%. The derived share rises from 27.4\% at 4k to 33.2\% at 64k and from 29.5\% at two hops to 40.8\% at four. Prompt reads recover increasingly sparse evidence, while \actTHINKop{} writes preserve intermediate results for reuse by later hops.

\paragraph{Memory addressing dominates the remaining multi-hop errors.}
When MuSiQue is split by hop count in \cref{tab:per-suite}, F1 falls from 42.1 at two hops to 16.4 at four, while recalls per position rise from 0.24 to 0.50. For wrong answers with three or more hops, a distractor outranks the supporting passage at the final evidence recall in 72\% of cases, while the supporting passage ranks second. The query $q$ scores the supporting key highly but fails to distinguish it from a nearby distractor.

\paragraph{RQ4: What does the write gate learn?}
We retrain the gate with the token-loss and write-time targets, record its value at each \actTHINKop{} step and whether a later positive-change recall addresses each write, and measure retention as the cosine between a memorized prompt value and its read-back after 256 to 2k generated tokens.

\paragraph{Gain credit opens the write gate for derived states that later recalls reuse.}
The write-time gate trains $g$ by whether \actTHINKop{} beats \actRECALL{} at the writing state. In \cref{tab:ablations}, these gates lower the long-context average by 1.6 and 2.1 points. Under gain credit, 58\% of reused writes have $g>0.7$, against 6\% of unread writes. \Cref{eq:write-credit} credits a write only when a later \actRECALL{} branch, executed or not, addresses its key and gains from its content, so redundant and unaddressed writes receive none.

\paragraph{A write-time target keeps the gate open and overwrites memorized evidence.}
In the third panel of \cref{fig:latent-dynamics}, the write-time gate remains high at every \actTHINKop{} step. In the fourth panel, its retention after 2k generated tokens falls to 0.64 against 0.91 under gain credit. Because the target is scored only where computing already beats recalling, the gate remains open for most writes regardless of whether a later recall reads them.

\section{Related Work}
\label{sec:related-work}

\paragraph{Memory for LLM reasoning.}
External memory systems~\citep{packer2023memgpt,xu2025mem,chhikara2025mem0,li2025memos,du2026memorysurvey} have progressed from preserving evidence beyond the model's active context to learning how stored information should be retained, retrieved, and transformed for subsequent reasoning. One line of work learns memory-management policies for writing, retaining, retrieving, and post-processing stored information~\citep{yu2025memagent,mem1_2025,wang2025mem,yuan2026memsearcher,membuilder2026,yu2026agentic,memoryr2_2026,zhang2026deltamem,li2026ember,ma2026memchain,dong2026memarbiter}. For example, Memory-R1~\citep{yan2025memory} learns a policy over memory actions through reinforcement learning, while MemRL~\citep{zhang2026memrl} applies runtime reinforcement learning to episodic memory. Mem-$\pi$~\citep{wang2026mempi} complements these approaches by learning when and what to write as generated memory. Another line of work changes either the representation or the use of stored evidence. It compresses or reorganizes the evidence itself, or connects retrieval to reasoning through interleaved search and long-context RL over extended inputs~\citep{li2025searcho1,song2025r1searcher,wan2025qwenlongl1,wu2025loongrl}. For example, R$^3$Mem~\citep{wang-etal-2025-r3mem} applies reversible compression, while Search-R1~\citep{jin2025searchr1} interleaves retrieval and reasoning over the extended input. These approaches manage stored information through external memory operations or extended token contexts. \method{} instead brings memory into recurrent latent computation, so recall and reasoning interleave within the same hidden-state trajectory and the latent action policy selects whether each state should recall missing evidence, continue computation, or exit.

\paragraph{Latent reasoning.}
Latent reasoning~\citep{zhu2025latentsurvey,latentcot2025survey} carries intermediate computation in hidden states and has progressed toward adapting both the form and the amount of computation to the current state. One line of work constructs and trains latent chains through pauses, hidden-state feedback, distillation, compression, adaptive routing, specialized memory tokens, and outcome-based objectives~\citep{goyal2023think,deng2023implicit,shen2025codi,xu2025softcot,zhang2025softthinking,tan2025think,simcot2025,softtokens2025,aichberger2026workingmemory,jung2026adaptivelatent,zou2026latentthoughtflow,zhao2026latentthoughtcredit}. For example, Coconut~\citep{hao2024training} feeds hidden states back as continuous thoughts, while System-1.5~\citep{system15_2025} routes between language and latent computation. Another line of work scales latent computation through recurrent depth and learns how to allocate or halt that computation~\citep{dehghani2019universal,geiping2025huginn,saunshi2025latentthoughts,haltinggates2026,schwethelm2026depthadaptive,lin2026allocating}. For example, Ouro~\citep{ouro2025} learns when to exit a weight-tied loop, while RLTT~\citep{rltt2026} assigns each recurrent state dense credit from the trajectory reward. Complementary work develops latent memory mechanisms~\citep{schlag2021fwp,yang2024deltanet,behrouz2024titans,behrouz2025atlas,kang2025lm2,memgen2025} that store and update hidden representations for later computation. \method{} differs by placing recurrent latent reasoning and fast-weight memory under one latent action policy over \actTHINKop{}, \actRECALL{}, and \actEXIT{}. \actTHINKop{} writes derived states to the same memory that stores input evidence, with the write gate credited by the gains of later recalls, while \actRECALL{} returns either distant evidence or prior latent computation to the loop.

\section{Conclusion}
\label{sec:conclusion}

We propose \method{}, which unifies memory access and reasoning through latent action selection over \actTHINKop{}, \actRECALL{}, and \actEXIT{} and one fast-weight memory of input evidence and derived states. Counterfactual policy distillation branches every latent action once to credit the policy, and gradients through the same branches credit the write gate. It outperforms token-space and latent-space baselines at lower inference compute, recalls where further computation cannot converge, and retains the derived states that later recalls reuse.

\bibliographystyle{plainnat}
\bibliography{references}

\end{document}